\documentclass[10pt,twocolumn,letterpaper]{article}

\usepackage[pagenumbers]{cvpr}

\usepackage{tabularx}
\usepackage{colortbl}

\AtEndPreamble{%
  \Crefname{figure}{Fig.}{Figs.}%
  \Crefname{table}{Tab.}{Tabs.}%
  \Crefname{section}{Sec.}{Secs.}%
  \Crefname{equation}{Eq.}{Eqs.}%
}

\definecolor{cvprblue}{rgb}{0.21,0.49,0.74}
\usepackage[pagebackref,breaklinks,colorlinks,allcolors=cvprblue,pdfpagemode=UseNone,bookmarksopen=false]{hyperref}

\title{\texorpdfstring{%
  \textcolor[RGB]{235,87,87}{J}%
  \textcolor[RGB]{39,174,96}{e}%
  \textcolor[RGB]{47,128,237}{v}eration%
}{Jeveration}: System One Models Now Generate}

\author{Yuhe Liu$^{*}$ \quad Qi Li$^{*}$ \quad Shihua Zhang \quad Xinchao Wang$^{\dagger}$\\
National University of Singapore\\
{\tt\small yuhe\_liu@u.nus.edu \quad liqi@u.nus.edu \quad suhzhang001@gmail.com \quad xinchao@nus.edu.sg}}

\makeatletter
\patchcmd{\@maketitle}{\vskip .375in}{\vskip 10pt}{}{\PackageError{arxiv}{the title block of cvpr.sty changed: spacing patch failed}{}}
\patchcmd{\@maketitle}{\vspace*{24pt}}{\vspace*{13pt}}{}{\PackageError{arxiv}{the title block of cvpr.sty changed: spacing patch failed}{}}
\patchcmd{\@maketitle}{\vspace*{12pt}}{\vspace*{-5pt}}{}{\PackageError{arxiv}{the title block of cvpr.sty changed: spacing patch failed}{}}
\makeatother

\begin{document}

\twocolumn[{\renewcommand\twocolumn[1][]{#1}\maketitle
  \begin{center}\captionsetup{hypcap=false}\noindent\includegraphics[width=\textwidth]{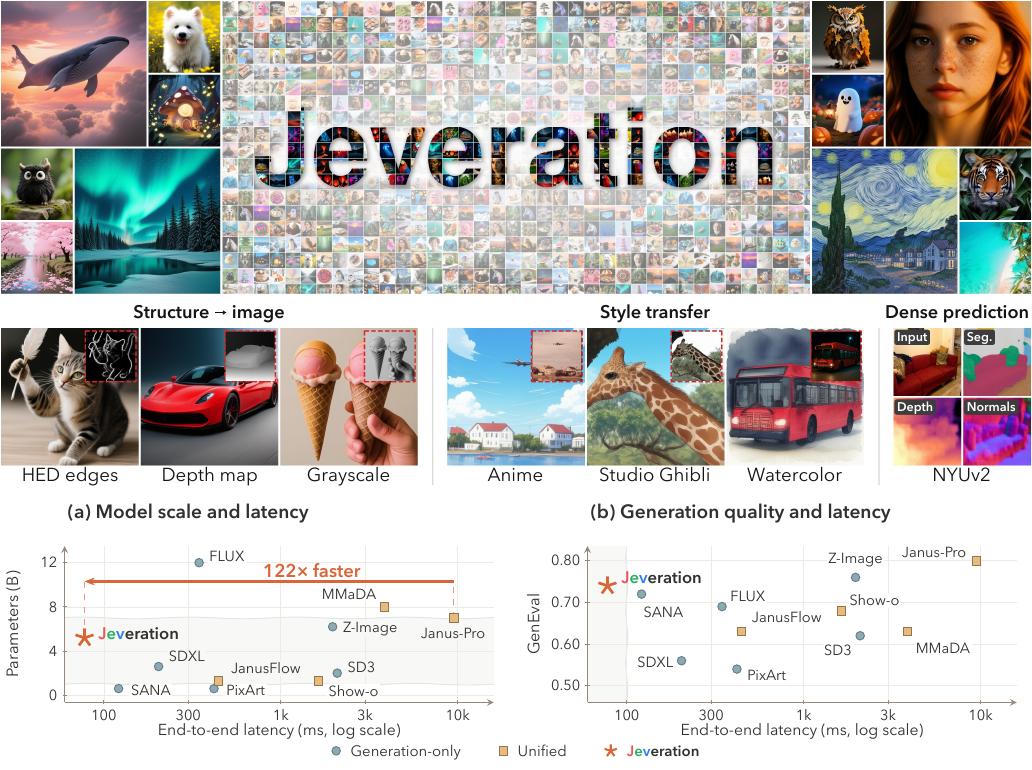}
  \captionof{figure}{\textbf{System One models now generate.} One frozen vision-language model regards text-to-image and image-to-image tasks as typed decisions. Text-to-image samples at 512\,px (top) and image-conditioned generation and dense prediction (middle), all from the same Jeveration checkpoint. The bottom charts plot model size (a) and GenEval (b) against latency on one H200.}\label{fig:teaser}\end{center}}]
{\renewcommand{\thefootnote}{\fnsymbol{footnote}}\footnotetext[1]{Equal contribution \quad \textsuperscript{\textdagger}Corresponding author}}
{\renewcommand{\thefootnote}{}\footnotetext{Project page: {\urlstyle{same}\url{https://carol-lyh.github.io/Jeveration/}}}}
{\renewcommand{\thefootnote}{}\footnotetext{Code: {\urlstyle{same}\url{https://github.com/Carol-lyh/Jeveration}}}}

\begin{abstract}
System One models such as Jev return decisions through fast, parallel prediction. In this paper, we introduce Jeveration, showing that the same decision paradigm can also generate high-quality images in a single forward pass. Jeveration builds on a frozen vision-language model, repurposing its visual-semantic knowledge for holistic image generation. The key challenge is to maintain global consistency. A single prompt admits many valid images, yet parallel predictions must converge on a coherent whole. We therefore condition all positions on a shared random state to coordinate parallel predictions, so that each draw yields a different but coherent image. Since the model tends to ignore this state under direct image supervision, we first learn a distribution over discrete image codes to guide single-pass generation. Beyond text-to-image generation, we further unify image-to-image generation and dense prediction within the same architecture, so that one checkpoint supports all three tasks, each in one forward pass. Experiments show that Jeveration generates a $512{\times}512$ image in 78\,ms, up to 122$\times$ faster than existing unified models, while maintaining competitive image quality and dense prediction performance.

\end{abstract}

\section{Introduction}
\label{sec:intro}

Image generation today is dominated by two paradigms. Autoregressive models~\cite{chameleon,emu3,januspro} generate an image token by token, while diffusion models~\cite{pixartalpha,sana,flux} generate it by iterative denoising. Both generally require many sequential model evaluations for a single image, which increases inference latency.
We introduce a third paradigm for image generation that removes sequential sampling, built on System One models. Our model, Jeveration, returns a complete image as one joint decision in a single forward pass, given a text prompt or a source image (\cref{fig:teaser}).

System One models such as Jev~\cite{jevdocs} return typed decisions, such as Choices or Scores, in a single parallel query without extended reasoning~\cite{jevdocs}. They have been applied to option selection, object localization, robot action scoring, and image evaluation~\cite{thisthat,jevbox,jevvla,visualjevrewards,li2026so}. Their effectiveness in these decision-making and understanding tasks draws in part on the visual and semantic knowledge acquired through large-scale pretraining. This knowledge also offers a foundation for image synthesis, suggesting that System One models have generative potential beyond their current uses. Motivated by this potential, we first explore a direct approach: using a System One model to select an entry from an existing codebook for every code position in a single forward pass. Our initial attempts, however, reveal two key challenges. First, direct parallel selection does not ensure global coherence. A single prompt admits many valid images, thus parallel decisions at different positions may correspond to incompatible realizations, producing image regions that fail to fit together, as illustrated in \cref{fig:mix}. Second, extending this approach to diverse image generation tasks requires a unified architecture that accommodates different inputs and conditioning requirements.

To keep the image globally consistent, we condition all positions on a shared random state that coordinates the parallel predictions, so that each draw yields a different but coherent image.
Furthermore, since a model trained on target images alone tends to ignore this state, we first train a model that learns the distribution of images as discrete codes, and then use it to judge the single-pass decisions and guide their training. The same paradigm extends to further tasks whose output is an image. Image-to-image generation uses a source image or visual conditions, such as edges, depth or grayscale, to guide the content and appearance of the generated image. Dense prediction can be regarded as image-to-image generation in a broad sense, where the model generates a map aligned with the input image that specifies a class, depth or surface normal at each pixel. One model can thus cover generation as well as perception.

Putting these together, Jeveration starts from a vision-language model (VLM), Qwen3.5-4B~\cite{qwen35}, and lets it answer visual tasks with Jev-style decisions. The VLM stays frozen, and Jeveration adds only LoRA adapters~\cite{lora} and output heads. Each answer, whether an image or a dense map, is one typed decision made in a single forward pass, and one checkpoint covers text-to-image generation, image-to-image generation and dense prediction tasks.
It reaches 0.74 GenEval at 512\,px in 78\,ms per image on one H200, 122$\times$ faster than Janus-Pro-7B~\cite{januspro}, closely follows the input structure in image-to-image generation, and achieves competitive results on dense prediction.
These results show that a decision model itself can generate images and predict dense maps in a single forward pass. Our contributions are:

\begin{figure}[t]
\centering
\raisebox{-30.87466pt}{\includegraphics[width=\linewidth]{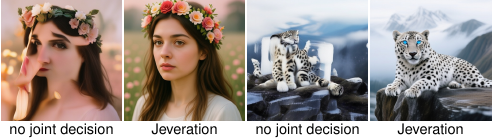}}
\caption{\textbf{Direct parallel selection yields incoherent images.} Codes sampled independently from the first-phase model do not fit together, while Jeveration produces coherent images.}
\vspace{-3mm}
\label{fig:mix}
\end{figure}
\begin{itemize}[leftmargin=*,nosep]
\item \textbf{Image generation through typed decisions.} To our knowledge, this is the first work to show that Jev-style system one models can themselves generate images in one forward pass, for both text-to-image and image-to-image generation.
\item \textbf{Learning coherent image decisions.} We let parallel decisions form one coherent image by conditioning them on one shared random state, and train the model to use this state by first learning the image distribution and then using it to judge and guide the decisions.
\item \textbf{One checkpoint across visual outputs.} One checkpoint with shared adapters performs text-to-image generation, image-to-image generation and dense prediction. Extensive experiments and ablation studies on these tasks demonstrate its effectiveness and efficiency.
\end{itemize}

\section{Related work}
\label{sec:related}

\paragraph{Unified multimodal understanding and generation.}
Unified multimodal models combine image understanding and synthesis. Chameleon~\cite{chameleon}, Emu3~\cite{emu3}, Janus/Janus-Pro~\cite{janus,januspro} and Unified-IO~2~\cite{unifiedio2} generate image tokens autoregressively. Show-o~\cite{showo} uses masked prediction, while Transfusion~\cite{transfusion} and BAGEL~\cite{bagel} model continuous image latents. MetaQueries~\cite{metaquery} and BLIP3-o~\cite{blip3o} instead connect understanding representations to separate diffusion generators. All of these models generate an image over many sequential steps, token by token or by iterative denoising. Dense prediction is another route to visual outputs. For instance, LENS~\cite{lens} adds segmentation to a frozen multimodal model, while DenseMLLM~\cite{densemllm} supervises visual tokens for dense prediction. In contrast, Jeveration unifies image synthesis and dense prediction by formulating both as parallel typed decisions, using a single frozen VLM with shared adapters and task-specific output heads.\looseness=-1

\paragraph{One-step image generation.}
Single-pass generators learn to turn a random input into an image without iterative sampling. Consistency models~\cite{consistency} and MeanFlow~\cite{meanflow} learn direct generation in continuous spaces. DMD/DMD2~\cite{dmd,dmd2}, SANA-Sprint~\cite{sanasprint} and TwinFlow~\cite{twinflow} provide further approaches to one-step or few-step synthesis. Acceleration also applies to unified models. For example, Show-o Turbo~\cite{showoturbo} uses consistency distillation, and Omni-Diffusion-Distill~\cite{omnidistill} compresses multimodal decoding trajectories. For discrete images, MaskGIT~\cite{maskgit}, Muse~\cite{muse} and Meissonic~\cite{meissonic} learn masked-code conditionals and sample over multiple rounds. Di[M]O~\cite{dimo} distills such a model into a one-step generator by matching its token-level conditionals with the help of an auxiliary model. Soft-Di[M]O~\cite{softdimo} further uses soft embeddings to enable differentiable refinement. 
Building on this line of work, Jeveration adapts token-level conditional matching to learn coherent image synthesis through the parallel typed decisions of a frozen VLM. Its target conditionals are provided by a masked model first trained on the same frozen backbone, while a confidence-guided correction target further guides parallel code selection.

\paragraph{Typed decision models.}
Jev~\cite{jevdocs} returns typed outputs such as Choices, Scores, and probabilities in parallel without generating lengthy text, while its Box interface predicts the four edges of an object's bounding box~\cite{jevbox}. Open decision models~\cite{thisthat,innerjev,ndm} and a clinical decision model~\cite{cs1} similarly provide structured decision outputs. Jeva~\cite{jeva} explicitly addresses candidate-set scalability, demonstrating full-set decision-making with up to 2,048 candidates in its evaluation. These decision interfaces also support specialized applications: Qwen-Audio-Jev~\cite{audiojev} answers typed questions about audio, JevVLA~\cite{jevvla} scores and corrects robot actions, and Visual Jev Rewards~\cite{visualjevrewards} evaluates generated images to reward a separate generator. Jeveration extends this paradigm to spatially coordinated image generation, predicting all image tokens in parallel within a single forward pass conditioned on a shared random state. The same model also predicts dense maps in one pass, using a class Choice at each position for segmentation and Scores for depth and surface normal estimation.

\section{Method}
\label{sec:method}

\begin{figure*}[t]
\centering
\includegraphics[width=\linewidth]{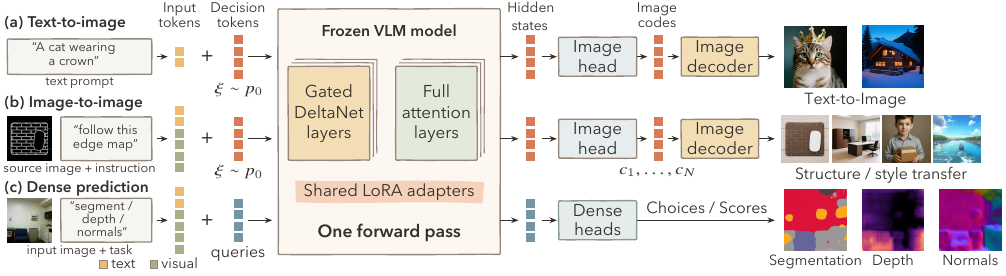}
\caption{\textbf{Architecture overview.} A frozen vision-language model with shared LoRA adapters processes input and decision tokens in one forward pass. Decision tokens start from a random state for image generation, while for dense prediction they act as spatial queries. Task-specific heads then map their final hidden states to image codes or dense predictions. One checkpoint serves all tasks.}
\label{fig:arch}
\end{figure*}

Jeveration enables a VLM to generate images through parallel typed decisions in a single forward pass, while supporting dense prediction within the same framework. As shown in \cref{fig:arch}, for image synthesis, these decisions select a codebook entry at each code position, while for dense prediction they produce task-specific classes or scores. A shared random state gives all code positions common stochastic conditioning, so that each draw yields a different but globally coherent image. We first formalize these visual decisions and the role of shared randomness (\cref{sec:readout}). We then describe how a frozen VLM with shared adapters and task-specific output heads supports these tasks within one checkpoint (\cref{sec:arch}). Finally, we present the training procedure that teaches the model to produce coherent images through parallel decisions (\cref{sec:training}).

\subsection{Images as Decisions}
\label{sec:readout}

Given a condition $y$, which holds a prompt or an input image, Jeveration answers with typed decisions~\cite{jevdocs}. An image is a joint \emph{Choice} of discrete codes $c=(c_1,\ldots,c_N)$ of an image tokenizer, segmentation is a \emph{Choice} over semantic classes at each spatial position, and depth and surface normals are real-valued \emph{Scores} at every position. A dense map has one correct answer per input and is decided deterministically, whereas one prompt admits many valid images, so an image decision should be stochastic.

\paragraph{Why a joint decision is hard.}
Given the conditioning input $y$, such as a text prompt or a source image with instructions, a direct approach predicts a categorical distribution at each code position and samples the codes independently in parallel:
\begin{equation}
q_{\mathrm{ind}}(c\mid y)=\prod_{i=1}^{N}q_i(c_i\mid y).
\label{eq:independent}
\end{equation}
This factorisation cannot capture dependencies between codes. For a fixed condition $y$ and target distribution \mbox{$p(c\mid y)$}, its best forward-KL approximation satisfies
\begin{equation}
\min_{q_{\mathrm{ind}}}\mathrm{KL}(p\,\|\,q_{\mathrm{ind}})=\sum_{i=1}^{N}H(c_i\mid y)-H(c\mid y),
\label{eq:tc_gap}
\end{equation}
where $H$ denotes entropy under $p$ given $y$. The gap is the target's total correlation, which measures how strongly the codes of an image depend on each other. A coherent image needs this dependence, but an independent model matches only the marginals $p(c_i\mid y)$, so its codes may follow different images and fail to fit together (\cref{fig:mix}).

\paragraph{A shared random state.}
Jeveration draws one random state sequence $\xi$ per image and conditions every code position on this shared state:
\begin{equation}
q_\theta(c\mid y)=\mathbb{E}_{\xi\sim p_0}\!\left[\prod_{i=1}^{N}q_{\theta,i}(c_i\mid y,\xi)\right].
\label{eq:decision}
\end{equation}
The shared state $\xi$ provides a common source of randomness for all image positions. Although the codes are sampled independently given $(y,\xi)$, their shared dependence on $\xi$ allows the resulting image distribution to capture dependencies between positions. The model can therefore coordinate its code choices while deciding all positions in a single forward pass. What remains is to learn how to use $\xi$ to produce coherent images (\cref{sec:training}).

\subsection{Architecture and Inference}
\label{sec:arch}
\label{sec:inference}

\paragraph{Overall architecture.}
We implement these visual decisions by appending decision tokens to the input of a VLM, as shown in \cref{fig:arch}. Text tokens encode the prompt and task instruction, and a frozen vision tower encodes any input image into visual tokens. The decision tokens are initialised from a sampled random state for image generation and serve as spatial queries for dense prediction. A frozen Qwen3.5-4B~\cite{qwen35} backbone with shared LoRA adapters~\cite{lora} processes the input and decision tokens in a single forward pass. In its full-attention layers, decision tokens attend to each other bidirectionally while the input prefix remains causal. Finally, task-specific heads map the decision tokens' final hidden states to image codes or dense predictions. The same checkpoint is used for
every task.

\paragraph{Image generation.}
The image tokenizer is a frozen TA-TiTok~\cite{titok,tatitok} with $N=128$ codes per image. For image generation, we append $N$ decision tokens, and the $i$-th token decides the code $c_i$. The initial embeddings of these tokens encode the random state $\xi\sim p_0$, which masks each position with probability $\rho$, fills the remaining positions with uniformly random codes and adds Gaussian noise of scale $\sigma$ to all embeddings. Following Di[M]O~\cite{dimo}, we set $\rho=0.6$ and $\sigma=0.1$. A transformer head~\cite{dit} reads the final hidden states of these tokens together with $\xi$ and outputs a feature $u_i$ for each position. We predict each code from $u_i$ with two branches. The codebook branch $f_{\mathrm{cb}}$ projects $u_i$ to the
code space and applies a classifier initialised from the tokenizer's codebook.
The wide branch $f_{\mathrm{wide}}$, initialised at zero, maps $u_i$ directly to the code logits, giving the head more freedom to learn. The code distribution sums the two:
\begin{equation}
q_{\theta,i}(\cdot\mid y,\xi)=\operatorname{softmax}\big(f_{\mathrm{cb}}(u_i)+f_{\mathrm{wide}}(u_i)\big).
\label{eq:image_head}
\end{equation}
Both branches are trained and we conduct ablations in \cref{sec:ablation} to show the benefit of combining them. At inference, all codes are sampled in parallel as one joint decision.
 Finally, they are rendered by the tokenizer's decoder, which also reads the prompt and which we fine-tune for 512-px output. For image-to-image generation, a gated branch also feeds the source image to the decoder.

\paragraph{Dense prediction.}
For dense prediction, the decision tokens are spatial queries aligned with the visual tokens of the input image. Each query starts from a learned vector plus a projection of its visual token. Task-specific heads decode each query's final hidden state into the segmentation, depth or surface-normal predictions for its image region.

\begin{figure*}[t]
\centering
\raisebox{-159.04494pt}{\includegraphics[width=\linewidth]{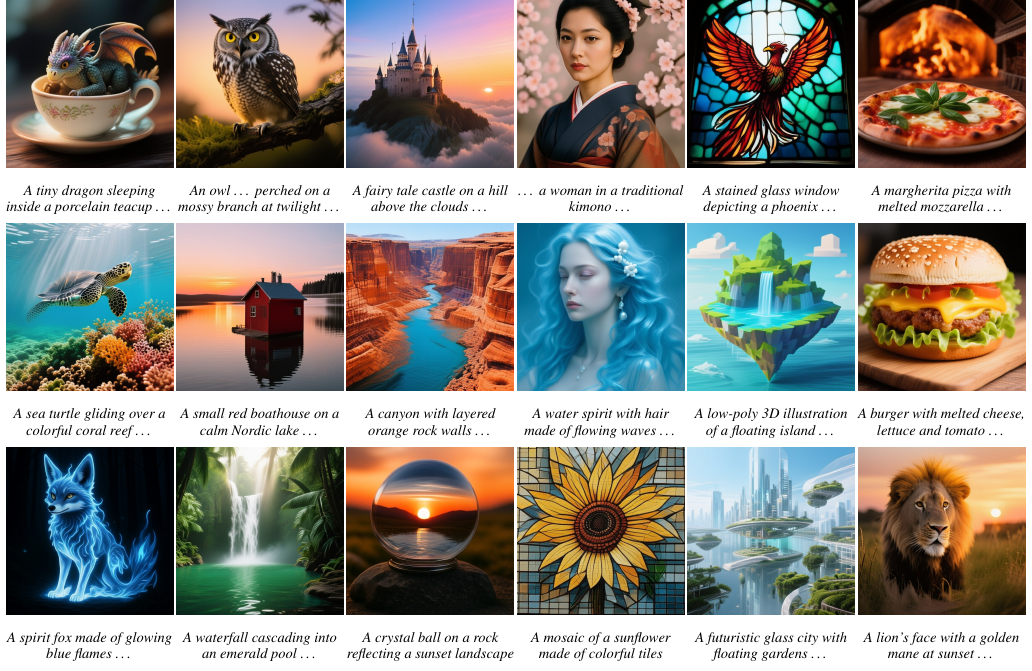}}
\caption{\textbf{Text-to-image generation.} Samples of the final Jeveration checkpoint at 512\,px, with excerpts of their prompts.}
\label{fig:t2i}
\end{figure*}

\subsection{Learning to Decide}
\label{sec:training}

We train the model to use $\xi$ to make coherent image decisions. Direct supervision on target codes does not encourage the model to use $\xi$, since the random state is sampled independently of the target image.
We therefore first learn a model of the image distribution through masked prediction, then use it to guide image decisions from random states.

\paragraph{Learning the image distribution.}
In the first phase, we train the architecture as a masked model~\cite{maskgit,muse}, keeping the VLM frozen. For target codes $c$, we mask a random set $M$ and predict the hidden codes from the visible codes $c_{\bar M}$ and condition $y$:
\begin{equation}
\mathcal L_{\mathrm{mask}}(\phi)=\mathbb E\!\left[\frac{1}{|M|}\sum_{i\in M}-\log p_{\phi,i}(c_i\mid c_{\bar M},y)\right].
\label{eq:teacher}
\end{equation}
The visible codes supply image-specific context, enabling the model to learn dependencies between positions. The resulting model provides reference conditional predictions for the second phase.

\paragraph{Learning image decisions.}
In the second phase, the decision model $\theta$ starts from the first-phase model and learns to turn random states into whole images. From $(y,\xi)$, $\theta$ predicts logits $\ell_{\theta,i}$ and samples a code sequence $\hat c$ in a single pass. We re-mask a random subset $M$ of $\hat c$, and two masked models predict the hidden codes from the same visible codes. The frozen first-phase model $\phi$ represents the image distribution and gives $p_{\phi,i}$. An auxiliary model $\psi$ gives $p_{\psi,i}$ for the current distribution of $\theta$. The auxiliary model begins as a copy of $\phi$ and keeps training on $\theta$'s samples with the masked loss of \cref{eq:teacher}, so it follows $\theta$ as $\theta$ changes. The two predictions agree when $\theta$ matches the image distribution. Following Di[M]O~\cite{dimo}, we use their difference to update $\theta$ by minimising
\begin{equation}
\mathcal L_{\mathrm{DM}}(\theta)=\mathbb E\!\left[\frac{1}{|M|}\sum_{i\in M}\mathrm{sg}(p_{\psi,i}-p_{\phi,i})^{\!\top}\ell_{\theta,i}\right],
\label{eq:dimo}
\end{equation}
where $\mathrm{sg}$ stops gradients.
 The prediction at each masked position depends on the other codes of the sample, so the two predictions agree only when $\theta$'s codes fit together. This pushes $\theta$ to coordinate its codes through the shared state $\xi$.

\begin{table*}[t]
\centering
\caption{\textbf{Text-to-image quality and end-to-end latency.} All models are timed on the same H200 at batch 1, including decoding, with released samplers at the listed steps and resolutions. Speedup is relative to Janus-Pro-7B, the slowest model listed. Params include our frozen backbone; Steps count sequential generator evaluations, or tokens for autoregressive models. Colored squares identify image-generation families, including distilled variants. $^\dagger$ / $^\ddagger$ indicate scores from the authors / another paper.}
\label{tab:main}
\footnotesize
\definecolor{tOneDiff}{HTML}{1F9ACB}
\definecolor{tOneMask}{HTML}{14966F}
\definecolor{tOneAR}{HTML}{AA66CC}
\definecolor{tOneDecision}{HTML}{E67E22}
\definecolor{tOneHeader}{HTML}{F0F3EA}
\definecolor{tOneOurs}{HTML}{FFF3E6}
\definecolor{tOneRule}{HTML}{000000}
\definecolor{tOneGroup}{HTML}{000000}
\newcommand{\tOneMark}[1]{\textcolor{#1}{\rule[0.1ex]{0.9ex}{0.9ex}}\hspace{0.45em}}
\setlength{\tabcolsep}{4.5pt}
\renewcommand{\arraystretch}{1.08}
\arrayrulecolor{tOneRule}
\resizebox{\textwidth}{!}{%
\begin{tabular}{lccccrcccr}
\toprule
\rowcolor{tOneHeader}
\textbf{Model} & \textbf{Unified} & \textbf{Params (B)} & \textbf{Steps} & \textbf{Res.\ (px)} & \textbf{Speedup}$\uparrow$ & \textbf{GenEval}$\uparrow$ & \textbf{DPG}$\uparrow$ & \textbf{HPSv2.1}$\uparrow$ & \textbf{Latency (ms)}$\downarrow$ \\
\midrule
\multicolumn{10}{l}{\textcolor{tOneGroup}{\emph{\textbf{Generation-only models}}}} \\
\tOneMark{tOneDiff}PixArt-$\alpha$~\cite{pixartalpha} & $\times$ & 0.6 & 20 & 512 & 25.1$\times$ & 0.49 & 71.6$^\ddagger$ & 29.98 & 378.8 \\
\tOneMark{tOneDiff}PixArt-$\Sigma$~\cite{pixartsigma} & $\times$ & 0.6 & 20 & 512 & 22.7$\times$ & 0.54 & 79.5$^\ddagger$ & 30.49 & 419.5 \\
\tOneMark{tOneDiff}SANA-0.6B~\cite{sana} & $\times$ & 0.6 & 20 & 512 & 16.2$\times$ & 0.65 & 84.3$^\dagger$ & 29.86 & 587.3 \\
\tOneMark{tOneDiff}SANA-Sprint 0.6B~\cite{sanasprint} & $\times$ & 0.6 & 1 & 1024 & 78.5$\times$ & 0.72 & 78.63 & 28.74 & 121.2 \\
\tOneMark{tOneDiff}SDXL-DMD2~\cite{dmd2} & $\times$ & 2.6 & 1 & 1024 & 46.6$\times$ & 0.56 & 74.09 & 28.20 & 204.0 \\
\tOneMark{tOneDiff}SDXL-LCM~\cite{lcm} & $\times$ & 2.6 & 4 & 1024 & 28.2$\times$ & 0.53$^\ddagger$ & 71.08 & 26.99 & 336.7 \\
\tOneMark{tOneDiff}FLUX.1-schnell~\cite{flux} & $\times$ & 12.0 & 1 & 1024 & 27.5$\times$ & 0.69 & 85.97 & 29.63 & 345.6 \\
\tOneMark{tOneDiff}Z-Image-Turbo~\cite{zimage} & $\times$ & 6.2 & 8 & 1024 & 4.8$\times$ & 0.76 & 84.99 & 29.76 & 1,965.2 \\
\tOneMark{tOneDiff}SD3-medium~\cite{sd3} & $\times$ & 2.0 & 28 & 1024 & 4.6$\times$ & 0.62$^\ddagger$ & 84.1$^\ddagger$ & 29.88 & 2,086.0 \\
\midrule
\multicolumn{10}{l}{\textcolor{tOneGroup}{\emph{\textbf{Unified models}}}} \\
\tOneMark{tOneDiff}JanusFlow-1.3B~\cite{janusflow} & \checkmark & 1.3 & 30 & 384 & 21.3$\times$ & 0.63$^\dagger$ & 80.09$^\dagger$ & 22.90 & 445.8 \\
\tOneMark{tOneAR}Janus-Pro-1B~\cite{januspro} & \checkmark & 1.5 & 576 & 384 & 1.3$\times$ & 0.73 & 82.63$^\dagger$ & 27.73 & 7,593.2 \\
\tOneMark{tOneAR}Janus-Pro-7B~\cite{januspro} & \checkmark & 7.0 & 576 & 384 & 1.0$\times$ & 0.80 & 83.91 & 28.47 & 9,508.4 \\
\tOneMark{tOneMask}Show-o Turbo-512~\cite{showoturbo} & \checkmark & 1.3 & 4 & 512 & 101.0$\times$ & 0.63 & 83.35 & 27.35 & 94.1 \\
\tOneMark{tOneMask}Show-o-512~\cite{showo} & \checkmark & 1.3 & 50 & 512 & 5.8$\times$ & 0.68 & 67.27$^\ddagger$ & 27.85 & 1,631.0 \\
\tOneMark{tOneMask}MMaDA-8B~\cite{mmada} & \checkmark & 8.0 & 15 & 512 & 2.5$\times$ & 0.63$^\dagger$ & 69.97$^\ddagger$ & 23.74 & 3,860.9 \\
\midrule
\multicolumn{10}{l}{\textcolor{tOneGroup}{\emph{\textbf{Ours}}}} \\
\rowcolor{tOneOurs}
\tOneMark{tOneDecision}\textbf{Jeveration} & \checkmark & 5.2 & 1 & 512 & \textbf{122.2$\times$} & 0.74 & 78.24 & 28.06 & 77.8 \\
\bottomrule
\end{tabular}}
\arrayrulecolor{black}
\par\vspace{3pt}
{\small
\tOneMark{tOneDiff}Continuous diffusion/flow family\quad
\tOneMark{tOneMask}Masked-token generation\quad
\tOneMark{tOneAR}Autoregressive generation\quad
\tOneMark{tOneDecision}Parallel typed decisions\par}
\vspace{-2pt}
\end{table*}

\paragraph{Confidence-guided correction.}
In the final part of the second phase, we add direct supervision on generated samples. For some samples in each batch, we re-mask the half of the positions with the lowest confidence $q_{\theta,i}(\hat c_i\mid y,\xi)$, denoted $R$, and let the frozen model complete them over several rounds into $c^{\mathrm{ref}}$. We then train $\theta$ to predict these corrected codes from the same $(y,\xi)$:
\begin{equation}
\begin{aligned}
\mathcal L_{\mathrm{gen}}&=\mathcal L_{\mathrm{DM}}+\lambda\,\mathcal L_{\mathrm{corr}},\\
\mathcal L_{\mathrm{corr}}&=\mathbb E_{\mathrm{selected}}\!\left[\frac{1}{|R|}\sum_{i\in R}-\log q_{\theta,i}(c_i^{\mathrm{ref}}\mid y,\xi)\right].
\end{aligned}
\label{eq:refine}
\end{equation}
The frozen model, the auxiliary model and the correction are used only in training, so inference remains a single forward pass of $\theta$.

\paragraph{Dense supervision.}
Dense tasks are supervised with ground truth, using cross-entropy for segmentation, an $\ell_1$ depth error and $1-\cos$ for surface normals.

\section{Experiments}
\label{sec:exp}

\subsection{Experimental Setup}
\label{sec:setup}

\paragraph{Data and training.}
All main results use the same Jeveration checkpoint. Generation training combines FLUX-Reason-6M images~\cite{fluxreason} and Qwen-Image~\cite{qwenimage} renderings of text-to-image-2M prompts~\cite{t2i2m}.
Image-to-image training uses derived conditions and synthetic style pairs, while dense tasks use NYUv2~\cite{nyuv2}. We train rank-256 LoRA adapters for about 92k steps in the first phase and 6.7k in the second, which take about four days on a single machine with four NVIDIA RTX PRO 6000 GPUs.

\paragraph{Quality evaluation.}
We evaluate text-to-image generation with GenEval~\cite{geneval}, DPG-Bench~\cite{dpg} and HPSv2.1~\cite{hpsv2}.
Image-to-image outputs are scored by how well they follow the input structure, using edge F1 or depth accuracy. Stylized outputs are compared with paired targets by PSNR. Dense prediction uses the standard NYUv2 metrics. Latency is the median of five end-to-end calls on one H200 at batch 1.
Additional details
are provided in the supplementary material.

\begin{figure*}[t]
\centering
\raisebox{-93.5552pt}{\includegraphics[width=\linewidth]{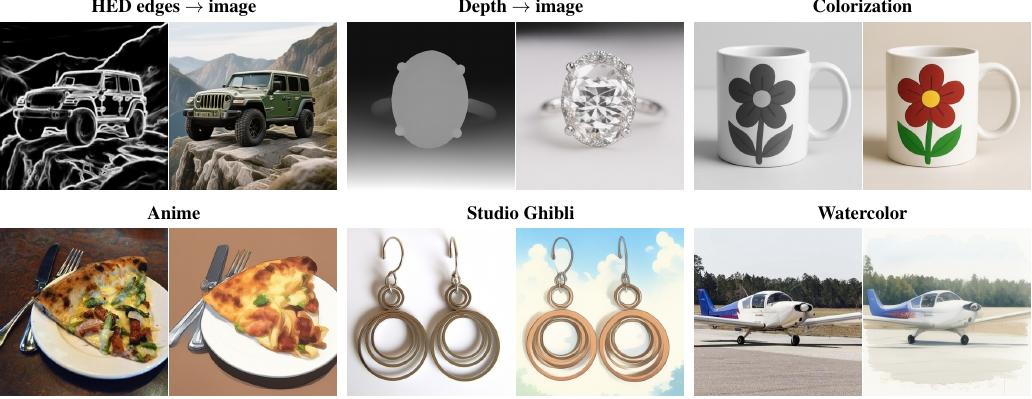}}
\caption{\textbf{Image-to-image generation} on held-out examples. In each pair the left image is the input and the right image the output.}
\label{fig:i2i}
\end{figure*}

\subsection{Text-to-Image Generation}
\label{sec:t2i}

Jeveration produces visually coherent images with detailed textures across natural scenes, portraits, objects and artistic styles (\cref{fig:teaser,fig:t2i}). The samples capture both the requested content and appearance, from the dragon inside a teacup to the stained-glass phoenix. In one forward pass, it generates a $512{\times}512$ image in 77.8\,ms end to end on one H200, the fastest in \cref{tab:main}, and reaches 0.74 GenEval.

\label{sec:speed}
Among unified models, Jeveration is the fastest and achieves higher GenEval and HPSv2.1 than JanusFlow, Show-o, Show-o Turbo, MMaDA and Janus-Pro-1B. At the same 512-px resolution, it improves GenEval by 0.11 over four-step Show-o Turbo-512 while reducing latency from 94.1 to 77.8\,ms. Compared with the autoregressive Janus-Pro-1B and Janus-Pro-7B, it is $98\times$ and $122\times$ faster.

Compared with generation-only models, Jeveration achieves higher GenEval than the multi-step models at the same 512-px resolution (0.74 versus 0.49 to 0.65) while being 5 to 8 times faster. It also achieves higher GenEval than SANA-Sprint 0.6B, FLUX.1-schnell and SDXL-DMD2, one-step models that generate 1024-px images. These results show that a frozen VLM can generate images of competitive quality in one forward pass.

\begin{table}[t]
\centering
\caption{\textbf{Image-to-image generation.} Structure fidelity on 341 held-out examples per task and style PSNR against paired targets.}
\label{tab:i2i}
\small

\definecolor{tOneHeader}{HTML}{F0F3EA}
\definecolor{tOneOurs}{HTML}{FFF3E6}

\setlength{\tabcolsep}{4pt}
\renewcommand{\arraystretch}{1.0}
\arrayrulecolor{black}

\begin{tabularx}{\columnwidth}{>{\raggedright\arraybackslash}Xcccc}
\toprule
 & \multicolumn{3}{c}{Structure follow$\uparrow$} & Style \\
\cmidrule(lr){2-4}\cmidrule(l){5-5}
\rowcolor{tOneHeader}
\textbf{Model} & \textbf{HED} & \textbf{Depth} & \textbf{Colorize} & \textbf{PSNR}$\uparrow$ \\
\midrule
\rowcolor{tOneOurs}
\textbf{Jeveration} & \textbf{0.880} & \textbf{0.802} & 0.937 & \textbf{15.40} \\
\midrule
Copy the input & 0.828 & 0.713 & \textbf{0.998} & -- \\
Decoder without source & 0.792 & 0.767 & 0.757 & 14.37 \\
\bottomrule
\end{tabularx}

\arrayrulecolor{black}
\end{table}

\begin{table}[t]
\centering
\caption{\textbf{Dense prediction on NYUv2.} Dense-only models use the same backbone. $^\ddagger$ is a reported result.}
\label{tab:dense}
\small

\definecolor{tOneHeader}{HTML}{F0F3EA}
\definecolor{tOneOurs}{HTML}{FFF3E6}

\setlength{\tabcolsep}{1.5pt}
\renewcommand{\arraystretch}{1.0}
\arrayrulecolor{black}

\begin{tabularx}{\columnwidth}{>{\raggedright\arraybackslash}Xcccc}
\toprule
 & Seg. & \multicolumn{2}{c}{Depth} & Normal \\
\cmidrule(lr){2-2}\cmidrule(lr){3-4}\cmidrule(l){5-5}
\rowcolor{tOneHeader}
\textbf{Model} & \textbf{mIoU}$\uparrow$ & \textbf{abs.\ (m)}$\downarrow$ & $\boldsymbol{\delta}_1\uparrow$ & \textbf{mean ($^\circ$)}$\downarrow$ \\
\midrule
\rowcolor{tOneOurs}
\textbf{Jeveration} & 67.7 & 0.267 & 0.898 & 24.2 \\
\midrule
Dense-only, one task each & \textbf{67.8} & \textbf{0.246} & \textbf{0.915} & \textbf{21.2} \\
Dense-only, three tasks & 67.3 & 0.253 & 0.912 & 23.0 \\
DeepLabV3+ R50$^\ddagger$~\cite{dbmtl,libmtl} & 53.5 & 0.393 & -- & 22.0 \\
\bottomrule
\end{tabularx}

\arrayrulecolor{black}
\end{table}

\subsection{Image-to-Image and Dense Prediction}
\label{sec:tasks}

\paragraph{Image-to-image generation.}
As shown in \cref{fig:i2i}, the same checkpoint handles many image-to-image tasks.
It can turn edge maps and depth maps into realistic images, colorize grayscale photos, and render photos in anime, Studio Ghibli and watercolor styles. Jeveration achieves strong structural fidelity, with 0.880 HED edge F1, 0.802 depth $\delta_1$ and 0.937 colorization edge F1 on held-out examples (\cref{tab:i2i}). Within this structure, the text prompt guides the content and appearance. Notably, this control comes at almost no extra cost, since each request is still a single forward pass of 91.7\,ms, only 14\,ms more than text-to-image generation. These results show that controllable generation is possible in one forward pass, without iterative sampling.

\begin{figure*}[t]
\centering
\raisebox{-67.54886pt}{\includegraphics[width=\linewidth]{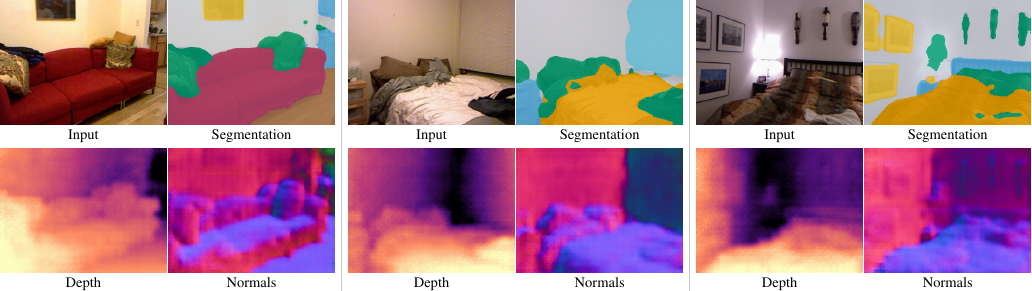}}
\caption{\textbf{Dense prediction} of 13-class segmentation, depth and surface normals on NYUv2 test pictures with the same checkpoint.}
\label{fig:dense}
\end{figure*}

\paragraph{Dense prediction.}
The same checkpoint is also a strong dense predictor (\cref{fig:dense}). As shown in \cref{tab:dense}, it reaches 67.7 mIoU on NYUv2, on par with dense-only models trained from the same backbone (67.8 and 67.3), and significantly outperforms DeepLabV3+, with 14.2 higher mIoU and a 32\% lower depth error (0.267 versus 0.393\,m). Each prediction takes a single forward pass of about 78\,ms. Surface normals lag behind, likely because fine local geometry is the most sensitive to sharing capacity across tasks. These results demonstrate that a single checkpoint can serve both image generation and dense prediction.

\vspace{5pt}
\subsection{Ablation Studies}
\label{sec:ablation}

\paragraph{Shared state and decision training.}
The shared random state lets all positions follow one image, while different states give different images. Decision training teaches the model to use this state. Without it, the first-phase masked model reaches only 0.32 GenEval in one pass (\cref{fig:mix}). With a fixed state instead of a random one, HPSv2.1 drops from 28.27 to 26.63 and GenEval from 0.73 to 0.70 (\cref{tab:abl_state}). The randomness of the shared state thus makes the images more coherent and of higher quality.

\begin{table}[htbp]
\centering
\caption{\textbf{Ablation studies.} Each block is a separate experiment that changes one component, so values are comparable only within a block. From top to bottom, the blocks start from a text-to-image-only model, an earlier run on all tasks and the first-phase masked model. Highlighted rows are the settings Jeveration uses.\looseness=-1}
\label{tab:abl_state}
\small

\definecolor{tOneHeader}{HTML}{F0F3EA}
\definecolor{tOneOurs}{HTML}{FFF3E6}

\setlength{\tabcolsep}{2pt}
\renewcommand{\arraystretch}{1.0}
\arrayrulecolor{black}

\begin{tabularx}{\columnwidth}{>{\raggedright\arraybackslash}Xcc}
\toprule
\rowcolor{tOneHeader}
\textbf{Variant} & \textbf{GenEval}$\uparrow$ & \textbf{HPS}$\uparrow$ \\
\midrule
\rowcolor{tOneOurs}
w/ random state
& \textbf{0.73} & \textbf{28.27} \\
w/ fixed state
& 0.70 & 26.63 \\
\midrule
\rowcolor{tOneOurs}
w/ correction target
& \textbf{0.74} & \textbf{28.03} \\
w/o correction target
& 0.73 & 27.92 \\
\midrule
\rowcolor{tOneOurs}
Image head (wide + codebook)
& \textbf{0.64} & \textbf{25.42} \\
Image head (wide only)
& 0.60 & 25.34 \\
Image head (codebook only)
& 0.59 & 25.04 \\
\bottomrule
\end{tabularx}

\arrayrulecolor{black}
\end{table}
\paragraph{Correction target.}
The correction target brings a further gain in image quality, as removing it lowers HPSv2.1 from 28.03 to 27.92 at similar GenEval (\cref{tab:abl_state}). This suggests that distribution matching already determines the composition, while the correction refines details.

\paragraph{Image-head branches.}
In the first-phase masked model, the two branches together outperform either branch alone, with 0.64 GenEval against 0.60 and 0.59 for the wide and codebook branches (\cref{tab:abl_state}). We hypothesize that they are complementary, as the codebook branch inherits the structure of the tokenizer's codebook while the wide branch adds capacity beyond this structure.

\paragraph{Source conditioning.}
Both the generated codes and the source image are essential for image-to-image generation. Even without the source, the decoder recovers most of the input structure from the codes alone (0.792 HED edge F1 and 0.767 depth $\delta_1$), indicating that the VLM transfers the input layout into its decisions. Feeding the source image to the decoder brings further improvements, raising HED edge F1 to 0.880 and colorization edge F1 from 0.757 to 0.937 (\cref{tab:i2i}). This suggests that the 128 codes capture the global structure, while the source image supplies the fine details.

\begin{table}[htbp]
\centering
\caption{\textbf{Joint training across tasks.} Training on all tasks costs almost no text-to-image quality. Both models get about the same amount of text-to-image training, so the difference comes from adding image-to-image and dense tasks.}
\label{tab:abl_unify}
\small

\definecolor{tOneHeader}{HTML}{F0F3EA}
\definecolor{tOneOurs}{HTML}{FFF3E6}

\renewcommand{\arraystretch}{1.0}
\arrayrulecolor{black}

\begin{tabularx}{\columnwidth}{>{\raggedright\arraybackslash}Xccc}
\toprule
\rowcolor{tOneHeader}
\textbf{Training tasks} & \textbf{GenEval}$\uparrow$ & \textbf{HPS}$\uparrow$ & \textbf{DPG}$\uparrow$ \\
\midrule
Text-to-image only & 0.73 & \textbf{28.28} & \textbf{78.61} \\
\rowcolor{tOneOurs}
All tasks (Jeveration) & \textbf{0.74} & 28.06 & 78.24 \\
\bottomrule
\end{tabularx}

\arrayrulecolor{black}
\end{table}
\paragraph{Joint training across tasks.}
Joint training extends the model to image-to-image generation and dense prediction at almost no cost to text-to-image quality. As shown in \cref{tab:abl_unify}, Jeveration performs on par with a model trained only on text-to-image, with comparable GenEval (0.74 versus 0.73), HPSv2.1 (28.06 versus 28.28) and DPG (78.24 versus 78.61). For dense prediction, it matches dense-only models in segmentation, with only a modest drop in depth and normal accuracy (\cref{tab:dense}). This demonstrates that a single set of adapters on a frozen VLM can accommodate diverse visual tasks with little interference.

\section{Conclusion}
\label{sec:limitations}
\label{sec:conclusion}

Jeveration introduces a third paradigm for image generation, in which a System One model returns a whole image as one joint decision in a single forward pass. Built on a frozen VLM, it conditions all code positions on a shared random state to encourage global consistency, while a masked model trained beforehand provides the target conditional distributions for parallel code prediction. One checkpoint performs text-to-image generation, image-to-image generation and dense prediction, and joint training across tasks comes at almost no cost to text-to-image quality. Jeveration generates 512-px images in 78\,ms on a single H200, achieves strong structural fidelity in image-to-image generation, and delivers segmentation performance comparable to dense-only models. We expect this paradigm to extend to further visual tasks and modalities.

{
    \small
    \bibliographystyle{ieeenat_fullname}
    \bibliography{jeveration}
}

\clearpage
\setcounter{page}{1}
\onecolumn
\begingroup
\setcounter{section}{0}
\setcounter{figure}{0}
\setcounter{table}{0}
\setcounter{equation}{0}
\setcounter{footnote}{0}
\renewcommand{\theHfigure}{supp.\arabic{figure}}
\renewcommand{\theHtable}{supp.\arabic{table}}
\renewcommand{\theHequation}{supp.\arabic{equation}}
\renewcommand{\theHsection}{supp.\arabic{section}}
\renewcommand{\thepage}{S\arabic{page}}
\captionsetup{hypcap=false}
\setlength{\parindent}{1em}
\raggedbottom
\begin{center}
{\Large\bfseries \thetitle\par}
\vspace{4pt}
{\large Supplementary Material\par}
\end{center}
\section{Natural Scenes and Objects}
\label{sec:supp_results}
All examples use the final Jeveration checkpoint, with one forward pass per task.
\begin{center}
\raisebox{-258.62708pt}{\includegraphics[width=\linewidth]{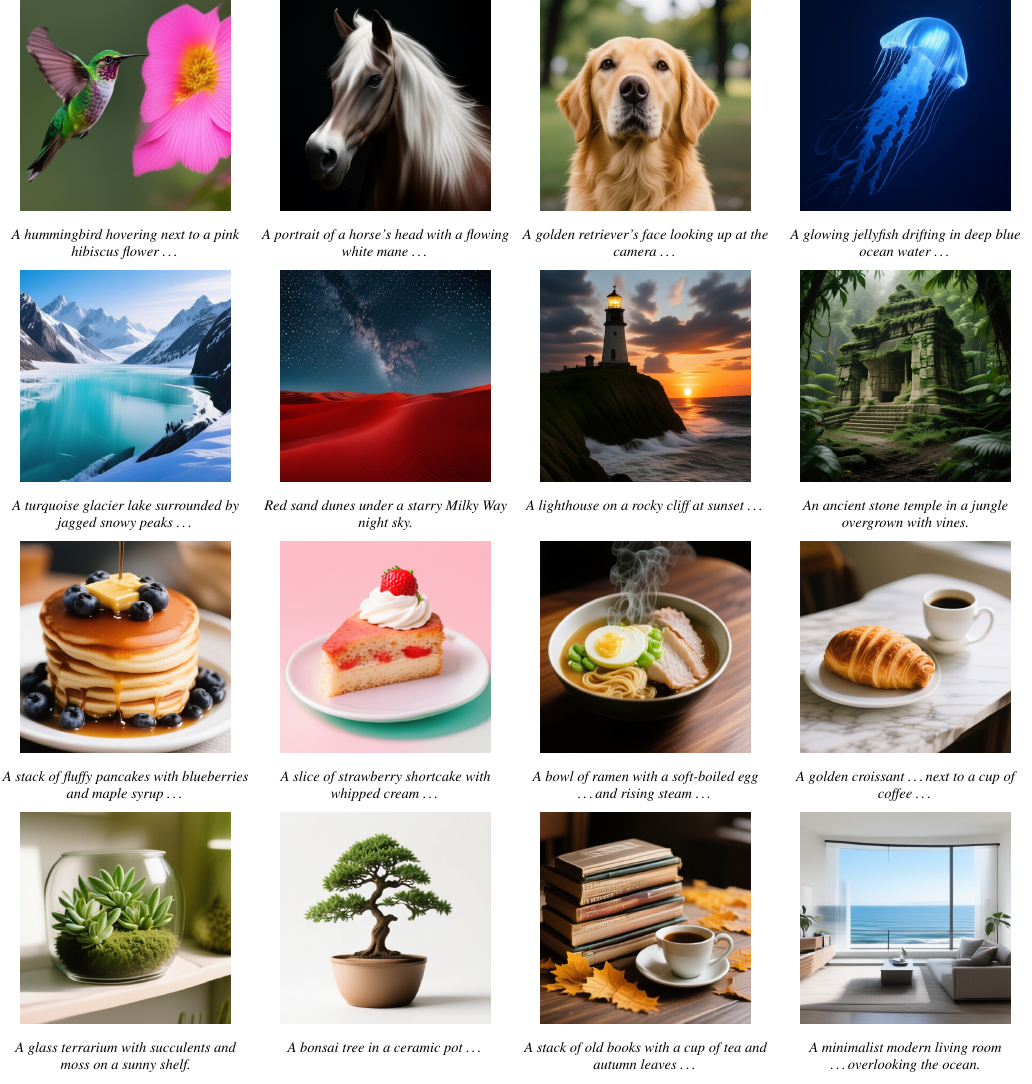}}
\captionof{figure}{\textbf{Additional text-to-image results.} Selected 512-px outputs across animals, landscapes, food and everyday scenes, with excerpts of their prompts.}
\label{fig:supp_t2i_natural}
\end{center}

\clearpage
\section{Artistic Styles and Portraits}
\begin{center}
\raisebox{-258.62708pt}{\includegraphics[width=\linewidth]{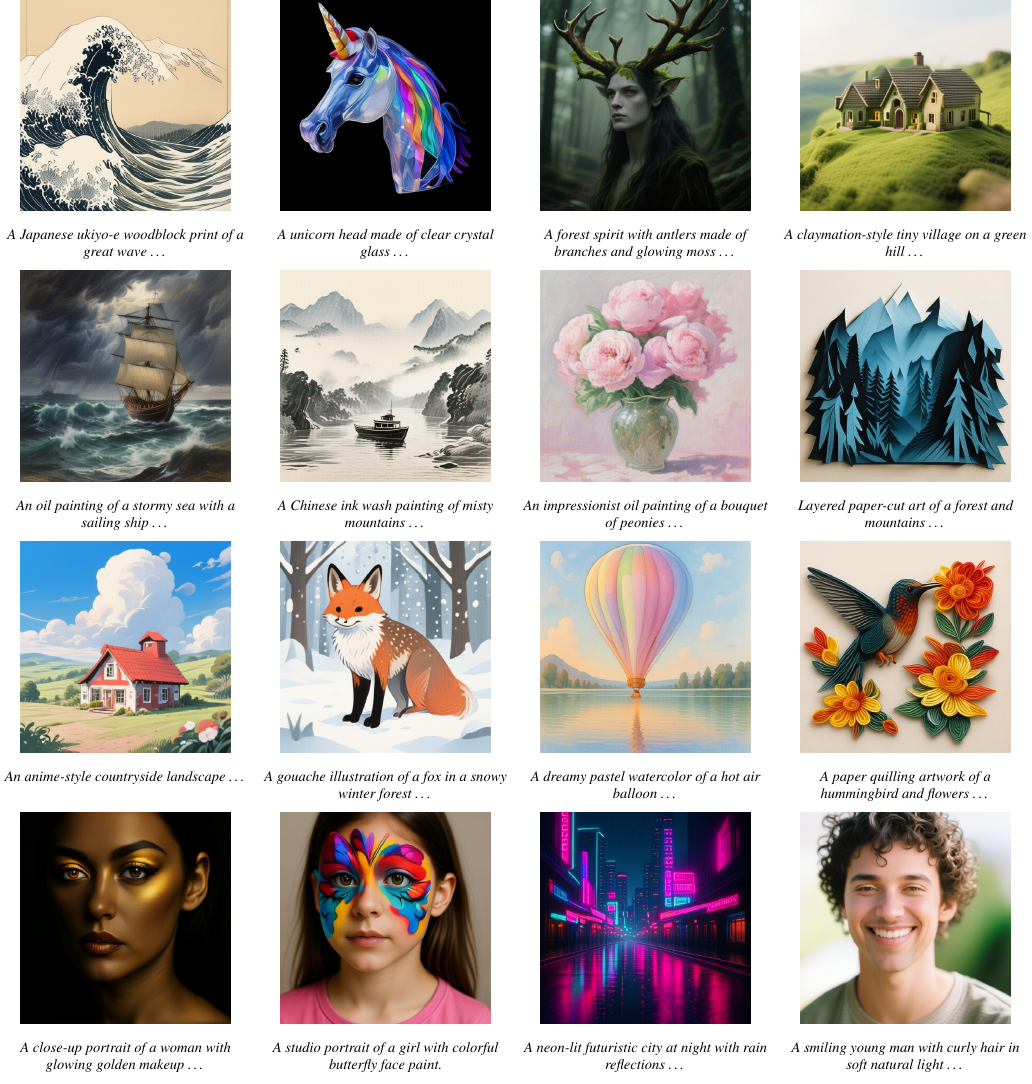}}
\captionof{figure}{\textbf{Fantasy, artistic styles and portraits.} Additional selected outputs of the same single-pass checkpoint.}
\label{fig:supp_t2i_styles}
\end{center}

\clearpage
\section{Image-to-Image Generation and Dense Prediction}
\begin{center}
\raisebox{-173.74463pt}{\includegraphics[width=\linewidth]{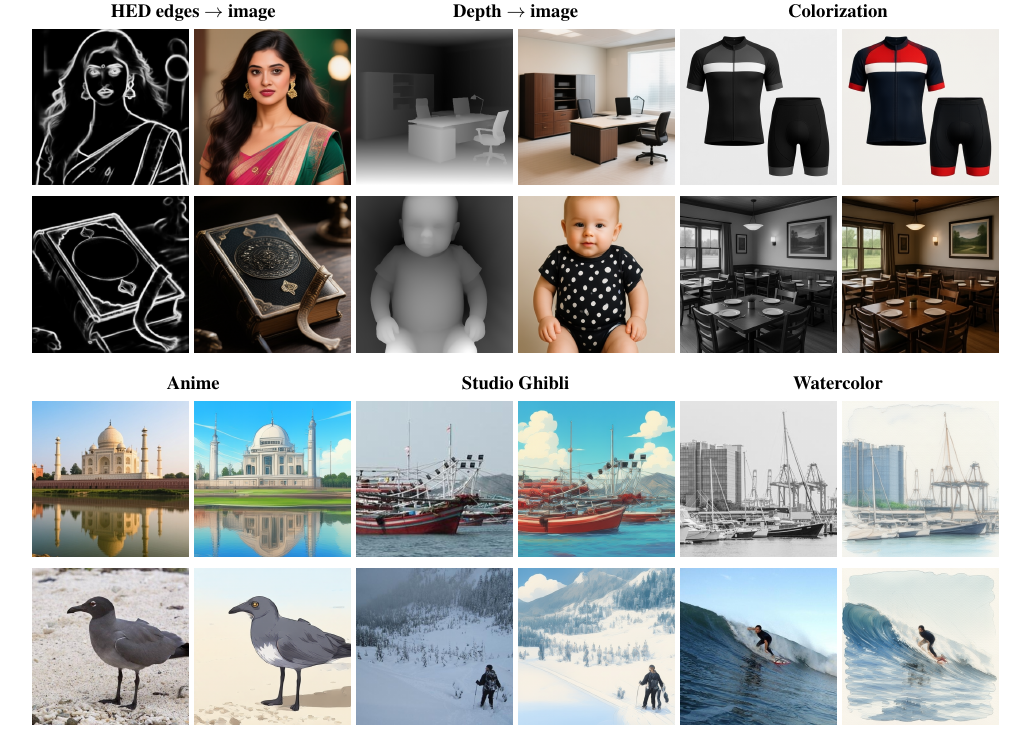}}
\captionof{figure}{\textbf{Additional image-to-image results.} Each pair shows the condition or source image on the left and Jeveration's output on the right. Structure tasks also receive a caption, while style tasks use a fixed instruction for each style.}
\label{fig:supp_i2i}
\end{center}
\begin{center}
\raisebox{-92.94432pt}{\includegraphics[width=\linewidth]{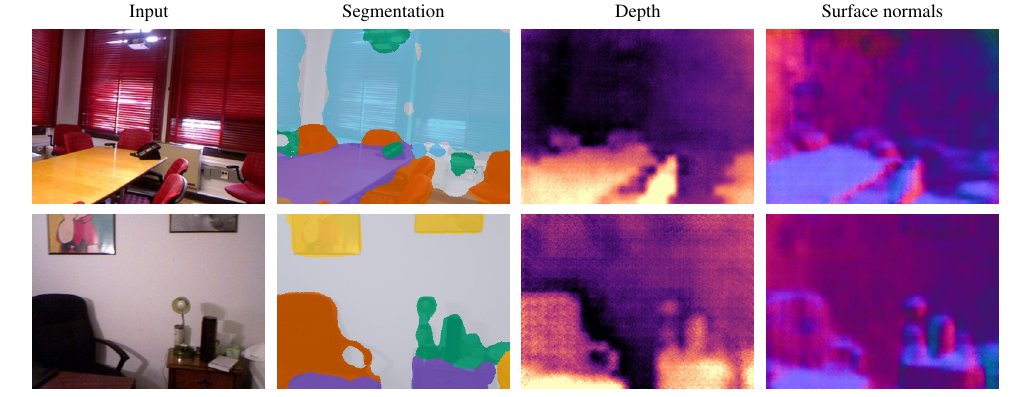}}
\captionof{figure}{\textbf{Dense outputs from the same checkpoint.} NYUv2 test images and single-pass predictions, arranged by task for comparison. Nearer depth is brighter.}
\label{fig:supp_dense}
\end{center}

\clearpage
\twocolumn
\section{Implementation and Evaluation}
\label{sec:supp_arch}
\label{sec:supp_hparams}
\label{sec:supp_dec}
\label{sec:supp_evaluation}
\label{sec:supp_latency}
\paragraph{Model.}
Jeveration keeps the Qwen3.5-4B backbone and its vision tower frozen. All tasks share rank-256 LoRA adapters, and task-specific heads read the decision tokens. Image generation uses 128 TA-TiTok codes from an 8,192-entry codebook, and dense prediction uses 432 spatial queries. About 1.0B parameters are trainable in total, including the adapters, input modules and output heads.

\paragraph{Training.}
Training has two phases, masked prediction followed by decision training with distribution matching and correction (\cref{tab:supp_hparams}). We use AdamW with an EMA of 0.999 on four NVIDIA RTX PRO 6000 GPUs. The correction re-masks the 50\% least confident positions of a sample, which the frozen first-phase model completes in four rounds at CFG 3, with $\lambda=1$.
\begin{center}
\captionof{table}{\textbf{Main training settings.} The backbone remains frozen in every stage.}
\label{tab:supp_hparams}
\small
\setlength{\tabcolsep}{2pt}
\begin{tabular*}{\columnwidth}{@{\extracolsep{\fill}}lrrr@{}}
\toprule
Training stage & Steps & LR & Batch \\
\midrule
Phase 1, generation only & 80k & $2\!\times\!10^{-4}$ & 256 \\
Phase 1, all tasks & 12k & $5\!\times\!10^{-5}$ & 256 \\
Phase 2, distribution matching & 5k & $5\!\times\!10^{-6}$ & 64 \\
Phase 2, with correction & 1.7k & $5\!\times\!10^{-6}$ & 64 \\
\bottomrule
\end{tabular*}
\end{center}

\paragraph{Decoding.}
We adapt the text-aware decoder of TA-TiTok to 512\,px and train it to reconstruct images from their codes. A source-conditioning branch is added for image-to-image tasks. The decoder relies on both of its inputs (\cref{tab:decoder}). Replacing the codes with those of another example lowers HED edge F1 from 0.896 to 0.552 and depth $\delta_1$ from 0.825 to 0.346, while replacing the source image lowers them to 0.652 and 0.767. This shows that the codes carry most of the structure and the source image refines it.

\begin{center}
\centering
\captionof{table}{\textbf{Code and source swap.} Each row replaces one decoder input with that of another example, on a shorter training run (32 examples per structure task, 223 style examples).}
\label{tab:decoder}
\small
\setlength{\tabcolsep}{2pt}
\begin{tabular*}{\columnwidth}{@{\extracolsep{\fill}}lccc@{}}
\toprule
Decoder input & HED$\uparrow$ & Depth$\uparrow$ & Style PSNR$\uparrow$ \\
\midrule
Correct codes and source & \textbf{0.896} & \textbf{0.825} & \textbf{15.43} \\
Wrong codes, correct source & 0.552 & 0.346 & 9.84 \\
Correct codes, wrong source & 0.652 & 0.767 & 14.71 \\
\bottomrule
\end{tabular*}
\end{center}

\paragraph{Evaluation.}
For text-to-image generation, GenEval uses its 553 prompts with four samples each, HPSv2.1 uses 3,200 prompts, and DPG-Bench scores 1,064 prompts as $2\!\times\!2$ grids. For image-to-image generation, structure fidelity is measured on 341 held-out examples per task. We extract the structure from each output and compare it with that of the reference image, using HED edge F1, depth $\delta_1$ after scale-and-shift alignment, and Canny edge F1 for colorization. Style PSNR compares each output with its paired target on 223 held-out examples. For dense prediction, we report the standard NYUv2 metrics on the 654 test images. The dense-only models use the same backbone and initialisation as Jeveration. All latencies are end-to-end times from prompt to decoded image on one H200 at batch 1, reported as the median of five calls after two warm-up calls.

\clearpage
\endgroup

\end{document}